\RequirePackage{fix-cm}
\documentclass[twocolumn]{article}          
\usepackage{graphicx}
\usepackage{url}
\usepackage{color}
\usepackage{amsfonts}
\begin{document}


\title{Determining the significance and relative importance of parameters of a simulated quenching algorithm using statistical tools
}

\author{P.A. Castillo \thanks{pedro@atc.ugr.es} \and M.G. Arenas \and N. Rico \and A.M. Mora \and \\ 
P. Garc\'{\i}a-S\'{a}nchez \and J.L.J. Laredo \and J.J. Merelo}


\date{} 

\maketitle

\begin{abstract}

When search methods are being designed it is very important to know which parameters have the greatest influence on the behaviour and performance of the algorithm.
To this end, algorithm parameters are commonly calibrated by means of either theoretic analysis or intensive experimentation. 
When undertaking a detailed statistical analysis of the influence of each parameter, the designer should pay attention mostly to the parameters that are statistically significant.
In this paper the ANOVA ({ANalysis Of the VAriance}) method is used to carry out an exhaustive analysis of a simulated annealing based method and the different parameters it requires.
Following this idea, the significance and relative importance of the parameters regarding the obtained results, as well as suitable values for each of these, were obtained using {ANOVA} and post-hoc Tukey HSD test, on four well known function optimization problems and the likelihood function that is used to estimate the parameters involved  in the lognormal diffusion process.
Through this statistical study we have verified the adequacy of parameter values available in the bibliography using parametric hypothesis tests. 
\end{abstract}

\section*{NOTE:}

This paper is a pre-print version of the paper below:\\

\textbf{Castillo, P.A., Arenas, M.G., Rico, N., Mora. A.M., Garcia P., Laredo, J.L.J., Merelo, J.J. Determining the significance and relative importance of parameters of a simulated quenching algorithm using statistical tools. Appl Intell 37, 239--254 (2012). https://doi.org/10.1007/s10489-011-0324-x}

\section{Introduction}
\label{intro}

When using search heuristics such as evolutionary algorithms (EAs)
\cite{Eiben2003,Michalewicz2004,yangXS2010}, simulated annealing (SA)
\cite{Kirkpatrick83,Ansari1993,adas2005,devicente2003} or local search
algorithms \cite{Hentenryck2005,Hoos2004,Bambha2004}, 
application rates for genetic operators, selection and replacement mechanisms, and the initial population, must first be chosen. 
The parameters used to apply these elements determine the way they operate and influence the results. 
Therefore, finding appropriate parameter values is a challenge in the field of metaheuristics, and can be addressed before the run (parameter tuning) or during the run (parameter control) \cite{Eiben99,Karr2003,SmitCEC2010}.

Due to the importance and effect of parameters on the results, a detailed statistical analysis of the influence of each parameter should be performed, paying attention mostly to those providing the values that are statistically significant.
In this paper, we propose using the ANOVA ({ANalysis Of the VAriance}) \cite{Fisher25}  statistical method as a tool to analyze a well known metaheuristic to solve function approximation problems.

This method allows us to determine whether a change in the results (responses) is due to a change in a parameter (variable or factor) or to a random effect, which makes it possible to determine the variables that have statistically significative effect on the method that is being evaluated.


The theory and methodology of ANOVA was mainly developed by R.A Fisher
during the 1920s \cite{Fisher25}. ANOVA examines the effects of one, two or more
quantitative or qualitative variables, called factors, on one quantitative
response. ANOVA is useful in a range of disciplines when it is suspected
that one or more factors might affect a response. It is essentially a
method used to analyze the variance to which a response is subjected,
dividing it into the various components corresponding to the sources of
variation, which can be identified. 


With ANOVA, we test a null hypothesis that all the population means are equal against the alternative hypothesis that there is at least two means that are not equal. 
We find the sample mean and variance for each level (value) of the main factor. The first one is obtained by finding the sample variance of the $n_k$ sample means from the overall mean. This variance is referred to as the \emph{variance among the means}. The second estimate of the population variance is found by using a weighted average of the sample variances. 
In order to solve this test, the significance level have been fixed in $\alpha=0.05$.



After applying ANOVA
, if the test shows significative differences, post-hoc tests
must be applied to find which levels are causing these differences. Tukey's
Honestly Significant Difference (HSD) \cite{nla.cat-vn1482421} has been used for this
purpose in this paper.

What we propose in this paper is a methodology based on applying the statistical parametric test ANOVA to determine the most important parameters of a simulated quenching (SQ) algorithm \cite{Ingber1993} (a SA-based optimization method) in terms of their influence on the results, and to establish the most suitable set of values for such parameters (thus obtaining an optimal operation).  

By using such a methodology, we might learn which parameters have almost no effect on the performance or which others have a strong correlation with fitness.
We think that our paper can help to find heuristic ways of obtaining the best result, or at least one that does not fall in a local minimum.
The main difference from previous approaches is that statistical tools to determine the significance and relative importance of parameters of a search method are used.

The rest of this paper is structured as follows:
Section \ref{sec:soa} presents a comprehensive review of the approaches found in the bibliography to describe different application parameters and determine the most suitable values for these parameters.
Section \ref{sec:method} briefly describes the SQ algorithm, and analyzes the parameters we propose to evaluate.
This section also presents the experimental setup and the methodology considered in the study. 
Section \ref{sec:Analysis} details the statistical study based on ANOVA. 
Obtained results are analyzed in order to establish the most suitable values.
Finally, conclusions and future work are presented in Section \ref{sec:conclusionsAndFutureWork}.

\section{Related work}
\label{sec:soa}

Adjusting the main design parameters has been usually solved either by hand \cite{Davis91,Jagielska99,Das2009} or using conventions, ad-hoc choices, intensive experimentation with different values \cite{DeJong75,emss2007,smiteiben_paper_2009,eiben_tut_cec2009,smiteiben_paper_cec2009,SmitCEC2010}, and even random initialization values \cite{GongCEC2011}, that is why new practices, based on well-grounded tuning methods (i.e. robust mathematical methods), are needed.
Such a methodology is what we propose in this paper.
As Eiben states, the straightforward approach is to generate and test \cite{Grefenstette86,DKim97,eiben_tut_cec2009,smiteiben_paper_cec2009}.
An alternative is to use a meta-algorithm to optimize the parameters \cite{Grefenstette86,Liang2001,Gacto2010,Diosan2010,Aksac2011}, that is, to run a higher level algorithm that searches for an optimal and general set of parameters to solve a wide range of optimization problems.
However, as some authors remark, solving specific problems requires specific parameter value sets \cite{Back97,Hinterding97,Eiben99} and, as Harik \cite{Harik99} claims, nobody knows the ``optimal'' parameter settings for an arbitrary real-world problem.
Therefore, establishing the optimal set of parameters for a sufficiently general case is a difficult problem.



Other researchers have proposed determining a good set of evolutionary algorithm parameters by analogy \cite{Back93,Goldberg89a,Goldberg91,Harik97,Harik99,Schaffer87,Thierens96}. 
In these works, a set of parameters is found, but instead of finding them by means of intense experimentation, the parameter settings are backed up with theoretical work (meaning that these settings are robust). 
Establishing parameters by analogy means using suitable sets of
parameters to solve similar problems; however they do not explain how
to establish the similarity between problems. 
Moreover, a clear protocol has not been proposed for situations when the similarity between problems implies that the most suitable sets of parameters are also similar \cite{Hinterding97,Eiben99}. 
Weyland has described a theoretical analysis of both an evolutionary  \cite{Weyland2007,Weyland2010} and simulated annealing algorithm \cite{Weyland2008} to search for the optimal parameter setting to solve the longest common subsequence problem. 
However, Weyland does not carry out this approach to practice.

Higher level algorithms \cite{Libro2006} have been proposed to face the problem of establishing parameter values:
several methods of testing and comparing different values before the run (parameter tuning), such as REVAC \cite{deSmitPPSN_11_Nannen2006,deSmitCEC_16_Nannen2007,deSmitPPSN_12_Smit2009}, SPOT \cite{deSmitPPSN_3_bartz2004,deSmitPPSN_4_bartz2004b,deSmitPPSN_2_bartz2005}, F-Race \cite{deMonteroCEC_4_birattari2002} and ParamILS \cite{deMonteroCEC_9_Hutter2007} are available. 
In this way, Smit et al. \cite{SmitCEC2010} propose improving the CEC2005 EA winner using REVAC as parameter tuning method.

Practical approaches based on setting parameter values on-line (during the run) instead of a previous parameter tuning have been proposed:
Montero et al. \cite{MonteroCEC2010} propose using off-line calibration techniques to detect whether a genetic operator help the evolutionary algorithm to perform his work, while 
Leung et al. \cite{LeungCEC2010} propose a novel adaptive parameter control system to control the parameters of a history-driven EA in an automatic manner. Their method obtains similar results to other methods while sets two EA parameters automatically.
Other proposed methods in this line are based on coding the parameters into the individual's genome (self-adaptation of parameters) or using non-static parameter settings techniques (controlled by feedback from the search and optimization process) \cite{emss2007,smiteiben_paper_cec2009,eiben2010_evostar}.
However, control strategies also have parameters and there are indicators that good tuning works better than control \cite{eiben_tut_cec2009}.

Recently, there has been a growing interest for the use of different statistical techniques in algorithms' comparison \cite{He2003,Maudes2011}. In this sense, Shilane et al. \cite{Shilane2008} proposes a framework for statistical performance comparison of evolutionary computation algorithms.
Garc\'{\i}a et al. \cite{Garcia2009} presents an analysis of the behaviour of evolutionary algorithms in optimization problems using parametric and non-parametric statistical tests.
Finally, Rojas et al. \cite{RojasIEEEanova} study the effects of parameters involved in GA design by applying ANOVA, while Castillo et al. \cite{CastilloIEEETNN} use statistical tools to find accurate parameter values involved in the design of a neuro-genetic algorithm. Methodology presented in these papers, could be improved by means of post-hoc statistical tests.

We propose a complete methodology based on parametric statistical tools to analyze the importance of parameters. 
Our proposal could be helpful to practitioners in designing and analyzing their optimization methods.

\section{Experimental Setup}
\label{sec:method}


Proposed methodology has been tested on a generalized version of the SA algorithm generally called simulated quenching \cite{Ingber1993}. 
Next, a brief description of the proposed algorithm is presented, focusing on detailing the parameters, the methodology and the problems used.

\subsection{Proposed Metaheuristic and Parameters Analysis}
\label{elmetodo}

SA is a generic probabilistic metaheuristic method for solving global optimization problems. 
Subject to conditions on the cooling schedule, simulated annealing can be shown to converge asymptotically to the global optima of the fitness function \cite{Ingber1993,Ansari1993,EGTalbi2009,LockettCEC2011}. In practice, simulated annealing has been used effectively in several science and engineering problems \cite{Debuse1999,Sun2008,Bonev2010}. However, its sensitivity to the proposal distribution and the cooling schedule means that it is not a good fit for all optimization problems \cite{EGTalbi2009,LockettCEC2011}.

In a SA algorithm, a cost function (objective function $f$) to be minimized is defined. 
Subsequently, from an initial random solution, different solutions are derived and compared to the current solution. The fittest solution (least cost) is kept, but retaining a solution with a greater cost is allowed with a certain probability, that decreases according to a ''temperature'' value.
Specifically, the algorithm generates a new state from the old one by means of a change operator. 
If the new one is better, then it is accepted. However, if the new one is worse, the algorithm will accept it with an acceptation probability of $p_a = e^{ - \Delta f / T }$ , where $\Delta f$  is defined as the increment of the cost function and $T$ is the temperature control parameter. The temperature is decreased using the temperature reduction function $f_T$ when a predefined number of new states from the current one have been generated. The simplest decrement rule is $T_{n+1} = \alpha T_n$ , where $\alpha$ is a constant smaller than 1 (in the interval $(0, 1)$). This exponential cooling scheme was proposed by Kirkpatrick et al. in \cite{Kirkpatrick83} with $\alpha \in [0.8, 0.95]$.

SA strength relies in its ability to statistically deliver the system to a global optimum, although in some cases the convergence could only be obtained after an infinite number of iterations (as stated in \cite{EGTalbi2009,LockettCEC2011}). 
However, due to its sensitivity to the proposal distribution and the cooling schedule, other algorithms (i.e. SA-like methods such as SQ) are usually used to find the global optimum faster. 
Although it is a faster method, it has the drawback of not having a convergence proof associated \cite{Ingber1993}.

Thus, in this paper a SQ algorithm, proposed by Michalewicz in \cite{Michalewicz96,Michalewicz2004}, is used. 
Additionally, our implementation uses several states at the same time (a population of individuals) instead of an isolated state, as proposed by Wang et al. in \cite{citaSAconPoblacion}.

Several parameters have been identified to configure the proposed optimization method. 
Our aim is to determine which parameters influence the obtained fitness and 
to facilitate practitioners some rules to decide which parameters to tune as they influence results in a different manner:
\begin{itemize}

        \item \textbf{Number of Changes} (\emph{NC}): this parameter represents the number of times that the cooling schedule is applied. This parameter controls the algorithm stopping criterion even if the temperature is not equal to 0. It sets how many times the cooling schedule is applied during the algorithm execution.

        \item \textbf{Population Size} (\emph{PS}): represents the number of states the algorithm uses instead of a single one as proposed by Wang et al in \cite{citaSAconPoblacion}.

        \item \textbf{Number of Iterations} (\emph{NI}): this parameter represents how many times the algorithm generates new neighbours to replace the current state.

        \item \textbf{Initial Temperature} (\emph{IT}): this parameter establishes the thermodynamic energy of the system. At the beginning this energy is high and it decreases with the cooling schedule \cite{EGTalbi2009}. 
        The temperature value controls the selective pressure: if the temperature is low, the probability of accepting a worse solution is low; while if temperature is high, the probability of accepting a worse solution is high, reducing the selection pressure this way.

        \item \textbf{Cooling Schedule} (\emph{CS}): this parameter represents how the temperature is gradually reduced as the simulation proceeds. Initially, temperature is set to a high value, and it decreases at each step according to the cooling schedule, that must be specified by the user \cite{EGTalbi2009}. At the end of the simulation process, the temperature will be close to 0. The system is expected to wander initially towards a broad region of the search space containing good solutions, ignoring small features of the energy function; then drift towards low-energy regions that become increasingly narrower; and finally move downhill according to the steepest descent heuristic.

\begin{table*}[htp]
\caption{\scriptsize{Parameters (factors) and the abbreviation used as reference later that determine the SA behaviour and values considered to apply ANOVA.}\label{tabla:parameters:sa}}
\begin{tabular}{|l|l|l|l|l|}
\hline
\scriptsize{Cooling}  &  \scriptsize{Number of}  &  \scriptsize{Number of}   & \scriptsize{Population} &  \scriptsize{Initial}      \\
\scriptsize{Schedule (CS)} &  \scriptsize{Changes (NC)}   &  \scriptsize{Iterations (NI)}  &   \scriptsize{Size (PS)}     &  \scriptsize{Temperature (IT)}  \\
\hline
\scriptsize{\textbf{C}auchy}          & \scriptsize{1000} \ \ \   & \scriptsize{2} \ \ \   & \scriptsize{1} \ \ \    &  \scriptsize{10}  \\
\scriptsize{Cauchy \textbf{M}odif.}   & \scriptsize{2000} \ \ \   & \scriptsize{4} \ \ \   & \scriptsize{2} \ \ \    &  \scriptsize{50} \\
\scriptsize{\textbf{E}xponential}     & \scriptsize{4000} \ \ \   & \scriptsize{8} \ \ \   & \scriptsize{4} \ \ \    &  \scriptsize{100} \\
                         & \scriptsize{8000} \ \ \   & \scriptsize{16} \ \ \  & \scriptsize{8} \ \ \    &  \\
                         & \scriptsize{16000} \ \ \  & \scriptsize{32} \ \ \  & \scriptsize{16} \ \ \   &  \\
\hline
\end{tabular}
\end{table*}

This parameter has an essential role in the efficiency and the effectiveness of the algorithm.
Three Cooling Schedules (those widely used by practitioners  \cite{citaTiposCS7,citaTiposCS2,Michalewicz2004,Michalewicz96,EGTalbi2009}) are used to determine which one yields better results. 
Using these schemes it can be seen as a way to speed the execution (i.e. using a Cauchy distribution makes the annealing schedule exponentially faster).

The \emph{Cauchy} cooling schedule follows  Equation \ref{Eq:cauchy} \cite{citaTiposCS7}:
\begin{equation}
T_{k} =  \frac{1}{1+k} 
\label{Eq:cauchy}
\end{equation} 
\noindent where $T_{k}$ is the temperature at iteration $k$.


The \emph{Modified Cauchy} cooling schedule follows  Equation \ref{Eq:cauchyMod} \cite{citaTiposCS7}:
\begin{equation}
\beta = \frac{T_{0}-T_{n}}{NC  T_{0} T_{n}} ;
T_{k} =  \frac{T_{k-1}}{1+\beta  T_{k-1}} 
\label{Eq:cauchyMod}
\end{equation} 
\noindent where $T_{k}$ is the temperature at iteration $k$, $T_{0}$ is the initial temperature, $T_{n}$ is the final temperature (fixed to $0.000001$) and $NC$ is the number of times we have to apply the cooling schedule.

The \emph{Exponential} cooling schedule follows  Equation \ref{Eq:exponencial} \cite{citaTiposCS2}:
\begin{equation}
T_{k} =  \alpha T_{k-1} 
\label{Eq:exponencial}
\end{equation} 
\noindent where $T_{k}$ is temperature at iteration $k$ and $\alpha$ usually takes a value between $0.85$ and $0.99$ (in these experiments, it is fixed to $0.9$).

\end{itemize}

As can be seen, this algorithm needs some parameters to be given adequate values. In literature, different sets of values can be found, obtained either using trial and error or by means of a theoretical analysis \cite{w14,Weyland2008}. 
However, our interest focuses on testing the effectiveness of those values using robust mathematical methods.

\subsection{Methodology}
\label{metodologia}

Once the optimization algorithm under study has been presented and the configuration parameters have been determined, the different levels (values) for those parameters have to be set before ANOVA is applied.

The set of values for each parameter was chosen taking into account those that can be found in the bibliography \cite{citaSAconPoblacion,citaTiposCS2,citaTiposCS5,citaTiposCS7,Michalewicz2004,EGTalbi2009}. 
Table \ref{tabla:parameters:sa} shows the different levels considered to evaluate these parameters using ANOVA. 
Values for \emph{NC}, \emph{NI} and \emph{IT} parameters were set incrementally.
In this case, 33750 runs were carried out for each problem (30 times * 3 levels of \emph{CS} * 5 levels for \emph{NC} * 5 levels for \emph{NI} * 5 levels for \emph{PS} * 3 levels for \emph{IT}, that represent the possible combinations) to obtain the fitness for each combination.


The application of ANOVA consisted in running the proposed SQ optimization method using those parameter combinations to obtain the best fitness. 
The response variable used to perform the statistical analysis is the fitness at the end of the simulation. The changes in the response variable are produced when a new combination of parameters is considered.
Then the {R} \footnote{{\tt http://www.r-project.org}} tool was used to obtain the ANOVA tables as well as the tables of means and figures for each problem. 
Appendix I shows how to use {R} to analyze data to generate tables and figures related to the Ackley problem (as an example).
Obtained ANOVA tables (Section \ref{sec:Analysis}) show for each factor, the freedom degree (FD), the value of the statistical F (F value) and its associated p-value.
As previously stated, if the output is smaller than 0.05, then the factor effect is statistically significant at a $95\%$ confidence level (what indicates that different initial values of this parameter give significative differences on the fitness).

As a significant F-value tells only that the effects are not all equal (i.e., reject the null hypothesis), post-hoc tests might be used to determine which effects or outputs are significantly different from which other.
In that sense, one of the most widely used post-hoc test is Tukey's Honestly Significant Difference test \cite{nla.cat-vn1482421}.
Tukey HSD is a versatile, easily calculated technique that might be used after ANOVA and allows saying exactly where the significant differences are. However, it can only be used when the ANOVA found a significant effect. 
If the F-value for a factor turns out non-significant, the further analysis is not needed.
The Tukey HSD compares pairs of the factor values, showing a segment (confidence level) for each comparison. 
Additionally, it shows a vertical dotted line (distance equals zero) that intersects some segments. Significant differences can be found in those cases where the vertical line does not intersect a segment (those values compared are significantly different).


\subsection{Function Approximation Problems}
\label{subsectionProblems}

In order to evaluate the proposed metaheuristic and their parameters, two experiments are proposed: in the first one (Section \ref{exp1}), four well known function approximation problems are used (Griewangk \cite{GRIE81}, Rastrigin \cite{TZ89}, Ackley \cite{funcs_ACK87,funcs_BAC96} and Rana \cite{funcs_Whitley03quadsearch}).
These problems are the more representative among those faced in our research.

In the second experiment (Section \ref{exp2}), the estimation of the parameters involved in a stochastic model (the lognormal diffusion process \cite{Gutierrez2001}) is carried out.

\medskip

Next, a detailed description of the four benchmark functions is given:

\begin{itemize}

\item The Griewangk function \cite{GRIE81} is a continuous multimodal function with a high number of local optima. It is defined according to Equation \ref{ecuacion:griewank} and its global optimum is located at point $x_{i}=0, i=1:100 $ \cite{DBLP:journals/amc/ChoOG08}. This problem has been addressed with vectors of n=100 real numbers in the interval $[-512, 512]$.
\begin{equation}
 f_{n}(\vec{x}) = \frac{1}{4000} \sum_{i=1}^n x_{i}^2 - \prod_{i=1}^n cos(\frac{x_{i}}{\sqrt{i}} + 1)
\label{ecuacion:griewank}
\end{equation}

\item The Rastrigin function \cite{TZ89} (see Equation \ref{ecuacion:rastrigin}) is a multimodal real function optimization problem,  whose global optimum is located at point $x_{i}=0, i=1:100 $ and whose minimum value is 0. This problem has been addressed with vectors of n=100 real numbers in the interval $[-50, 50]$.  
\begin{equation}
 f_{n}(\vec{x}) = 10 n +\sum_{i=1}^n x_{i}^2 - 10 cos(2 \pi x_{i}) \\
\label{ecuacion:rastrigin}
\end{equation}

\item The Ackley function \cite{funcs_ACK87,funcs_BAC96} (see Equation \ref{ecuacion:Ackley}) is a multimodal non separable and regular function usually used as test function. The global optimum is located at point $x_{i}=0, i=1:100 $. This problem has been addressed with vectors of n=100 real numbers in the interval $[-100, 100]$. 
\begin{equation}
f_{n}(\vec{x}) = 20+ e -20 \exp [-0.2\sqrt{\frac{1}{n} \cdotp \sum_{i=1}^n x_{i}^2 } ]
\label{ecuacion:Ackley} 
\end{equation}

\begin{center}

$ - \exp [\frac{1}{n} \sum_{i=1}^n (cos(2 \pi x_{i})] $

\end{center}

\item The Rana function \cite{funcs_Whitley03quadsearch} (see Equation \ref{ecuacion:Rana}) is a non-separable, highly multimodal function, whose global optimum is located at point $x_{i}=-514.04, i=1:100$. Vectors of n=100 real numbers in the interval $[-520, 520]$ have been considered.  
\begin{eqnarray}
f_{n}(\vec{x}) = \sum_{i=1}^n( (x_{i+1}+1) cos(\sqrt{\left | { x_{i+1}-x_{i}+1 }\right |})
\label{ecuacion:Rana} 
\end{eqnarray}

\begin{center}
 
$ sin(\sqrt{\left | {x_{i+1}+x_{i}+1  }\right |}) $

~\\

~\\

$ + x_{i} cos(\sqrt{\left | { x_{i+1}+x_{i}+1 }\right |}) 
 sin(\sqrt{\left | {x_{i+1}-x_{i}+1  }\right |})) $

 \end{center}

\end{itemize}

In all cases, as the optimum is known, the fitness of an individual is calculated as the distance to the optimum for that function, and the goal is to obtain the smallest fitness for the optimized function.

\medskip

On the second experiment, we propose estimating the parameters of lognormal diffusion process.
This stochastic process has been widely studied from the point of view of his applications for modeling real data. This is because there exist a lot of variables that are inherent positive, such as population size in biology \cite{Capocelli1974}, precipitations in meteorology, gas emission in environmental sciences, gross national product or consumer price index in economy \cite{Basel2004} or maternal age in the first pregnancy in demography. 
The use of a stochastic process for modeling these situations is due by the fact that deterministic models can not embrace all the factors that are involved in this kind of phenomena. 

The univariate homogeneous lognormal diffusion process is defined by a diffusion $\{X(t);t_0 \leq t \leq T\}$, 
taking values on $\mathbb{R}^+$ 
and with infinitesimal moments $A_1(x)=mx$ and $A_2(x) = \sigma^2 x^2$, 
where $m \in \mathbb{R} $ 
and $\sigma > 0$. 

\begin{table*}[htp]

\caption{\scriptsize{ANOVA tables for the fitness (response) with the parameters as factors. Note that in all functions a change in \emph{CS} parameter significantly influences the results. Factor \emph{PS} is not relevant in any case, which confirms that SQ algorithm operation is accurate using just one individual.}
\label{tablas:anova:simann}}

\begin{center}
\begin{tabular}{cccc}

\begin{tabular}{|c|c|c|c|}
\hline
\scriptsize{Param.}		&\scriptsize{FD}&  \scriptsize{F} & \scriptsize{$P (Sig. Level)$} \\ 
\hline
\scriptsize{\textbf{CS}}&\scriptsize{2}&\scriptsize{$  1592.9 $}&\scriptsize{\textbf{$ < 2.2e-16 $}}\\ 
\scriptsize{NC}          &\scriptsize{4}&\scriptsize{$ 0.5001  $}&\scriptsize{$ 0.7357 $}\\ 
\scriptsize{NI}          &\scriptsize{4}&\scriptsize{$ 0.0562  $}&\scriptsize{$ 0.9941 $}\\
\scriptsize{PS}          &\scriptsize{4}&\scriptsize{$  0.6748 $}&\scriptsize{$ 0.6095 $}\\
\scriptsize{\textbf{IT}} &\scriptsize{2}&\scriptsize{$ 53.111  $}&\scriptsize{\textbf{$< 2.2e-16  $}}\\ 
\hline
\end{tabular}

& & &

\begin{tabular}{|c|c|c|c|}
\hline
\scriptsize{Param.}		&\scriptsize{FD}&  \scriptsize{F} & \scriptsize{$P (Sig. Level)$} \\ 
\hline
\scriptsize{\textbf{CS}} &\scriptsize{2}&\scriptsize{$ 571.45  $}&\scriptsize{\textbf{$ < 2.2e-16 $}}\\ 
\scriptsize{\textbf{NC}} &\scriptsize{4}&\scriptsize{$ 14.699  $}&\scriptsize{\textbf{$ 1.042e-11 $}}\\ 
\scriptsize{\textbf{NI}} &\scriptsize{4}&\scriptsize{$ 8.2081  $}&\scriptsize{\textbf{$ 1.593e-06 $}}\\
\scriptsize{PS}          &\scriptsize{4}&\scriptsize{$ 0.9574  $}&\scriptsize{$ 0.43 $}\\
\scriptsize{\textbf{IT}} &\scriptsize{2}&\scriptsize{$  79.214 $}&\scriptsize{\textbf{$ < 2.2e-16 $}}\\ 
\hline
\end{tabular}

\\

 & & &  \\
\textbf{Griewangk} & & & \textbf{Rastrigin} \\
  & & &   \\
 & & &  \\

\begin{tabular}{|c|c|c|c|}
\hline
\scriptsize{Param.}		&\scriptsize{FD}&  \scriptsize{F} & \scriptsize{$P (Sig. Level)$} \\ 
\hline
\scriptsize{\textbf{CS}}    &\scriptsize{2}&\scriptsize{$ 378.38  $}&\scriptsize{\textbf{$ < 2.2e-16 $}}\\ 
\scriptsize{\textbf{NC}}    &\scriptsize{4}&\scriptsize{$ 45.951  $}&\scriptsize{\textbf{$ < 2.2e-16 $}}\\ 
\scriptsize{\textbf{NI}}    &\scriptsize{4}&\scriptsize{$ 7.6122  $}&\scriptsize{\textbf{$ 4.739e-06 $}}\\
\scriptsize{PS}             &\scriptsize{4}&\scriptsize{$ 0.1176  $}&\scriptsize{$ 0.9763 $}\\
\scriptsize{IT} 	        &\scriptsize{2}&\scriptsize{$ 0.0073  $}&\scriptsize{$ 0.9927 $}\\ 
\hline
\end{tabular}

& & &

\begin{tabular}{|c|c|c|c|}
\hline
\scriptsize{Param.}		&\scriptsize{FD}&  \scriptsize{F} & \scriptsize{$P (Sig. Level)$} \\ 
\hline
\scriptsize{\textbf{CS}} &\scriptsize{2}&\scriptsize{$ 2701.9  $}&\scriptsize{\textbf{$ < 2.2e-16 $}}\\ 
\scriptsize{\textbf{NC}} &\scriptsize{4}&\scriptsize{$ 10.244  $}&\scriptsize{\textbf{$ 3.788e-08 $}}\\ 
\scriptsize{NI}          &\scriptsize{4}&\scriptsize{$ 2.1146  $}&\scriptsize{$ 0.0769 $}\\
\scriptsize{PS}          &\scriptsize{4}&\scriptsize{$  0.5427 $}&\scriptsize{$0.7044  $}\\
\scriptsize{IT} 	     &\scriptsize{2}&\scriptsize{$ 2.273  $} &\scriptsize{$ 0.1035 $}\\ 
\hline
\end{tabular}

\\

 & & &  \\
\textbf{Ackley} & & & \textbf{Rana} 

\end{tabular}
\end{center}

\end{table*}

It has been applied for modeling phenomena with random variables that evolve through the time and show exponential trends. Moreover, its mean and mode functions could be used for predictions. Therefore, the inference on these two functions has been object of considerable study. 
In Guti\'errez et al. \cite{Gutierrez2001} a general study was made to obtain maximum likelihood estimators (MLE) and uniformly minimum variance unbiased estimators (UMVUE) for general parametric functions of the lognormal diffusion process, that are both based on the MLE of the parameters $a$ and $\sigma^2$. 
For fixed times $t_1, ..., t_n$, we observe the variables $X(t_1), ..., X(t_n)$ and their values $x_1, ..., x_n$ make up the basic sample from the  inferential process is carried out. Assuming $P[X(t_1)=x_1]=1$, we can get MLE for the parameters by maximizing the likelihood function, Equation \ref{ecuacion:rwp}, for the sample:


\begin{eqnarray}
L(a,\sigma^2)=\frac{1}{(2\pi)^{(n-1)/2}(\sigma^2)^{(n-1)/2}} 
\label{ecuacion:rwp} 
\end{eqnarray}

\begin{center}
 
$ \exp (- \frac{1}{2\sigma^2}\sum_{i=2}^n ( \frac{(\ln(x_i)-\ln(x_{i-1}))^2}{t_i-t_{i-1}} $

~\\

~\\

$ + a^2(t_i-t_{i-1}) -2a (\ln(x_i)-\ln(x_{i-1})) ) ) $

 \end{center}


In this problem we propose searching for the optimum parameter values of the likelihood function  using SQ in order to maximize it.
The aim is to establish which of the parameters have a significant effect on the estimate.
To do so, a path of simulated diffusion process with $m=0$, $t_i=i-1$, $i=1,...,101$ y $\sigma^2=10^{-5}$ is taken.

\medskip

Finally, paying attention to the change operator (or new state generator), and taking into account that individuals are vectors of real numbers, a uniform random mutation is used. 
A uniform random variable in the interval $[a,b]$ is generated. The parameter $a$ is in general equal to $-b$. The offspring is generated within the hyperbox $x+U(-b,b)^n$ where $b$ represents a constant that depends on the interval the real numbers are defined \cite{Herrera03ataxonomy,EGTalbi2009}.

\section{Statistical Study and Obtained Results} 
\label{sec:Analysis}

In this section, the ANOVA and Tukey HSD statistical tools are applied to determine whether the influence on parameter values (factors) is significant in the obtained fitness. 
ANOVA tables are shown as well as Tukey HSD plots and tables of means (showing the effect of each parameter level on the fitness).

As stated before, in the Subsection \ref{exp1} four well known function approximation problems are used, while Subsection \ref{exp2} analyses a much complex problem (the estimation of the parameters involved in the lognormal diffusion process).

\subsection{Experiment 1: Benchmark functions}  
\label{exp1}


Table \ref{tablas:anova:simann} shows the result of applying ANOVA on proposed approximation function problems using the SQ algorithm. 
Parameters with a significance level over $95\%$ are highlighted in boldface.

The ANOVA analysis shows that \emph{CS} parameter influence the obtained fitness, which indicates that changes in this parameter influence the results significantly. 
Note that \emph{PS} has not been reported as significant according to ANOVA tables, which confirms the fact that SA algorithm operation is accurate using just one individual.
Noteworthy that for some problems \emph{NC}, \emph{NI} and \emph{IT} are significant, though not in all cases.
This fact shows how for each problem different set of parameters can influence results in a different manner \cite{Wolpert1997}.

\begin{figure*}[htp]
\begin{center}
\includegraphics[width=6cm,height=5cm]{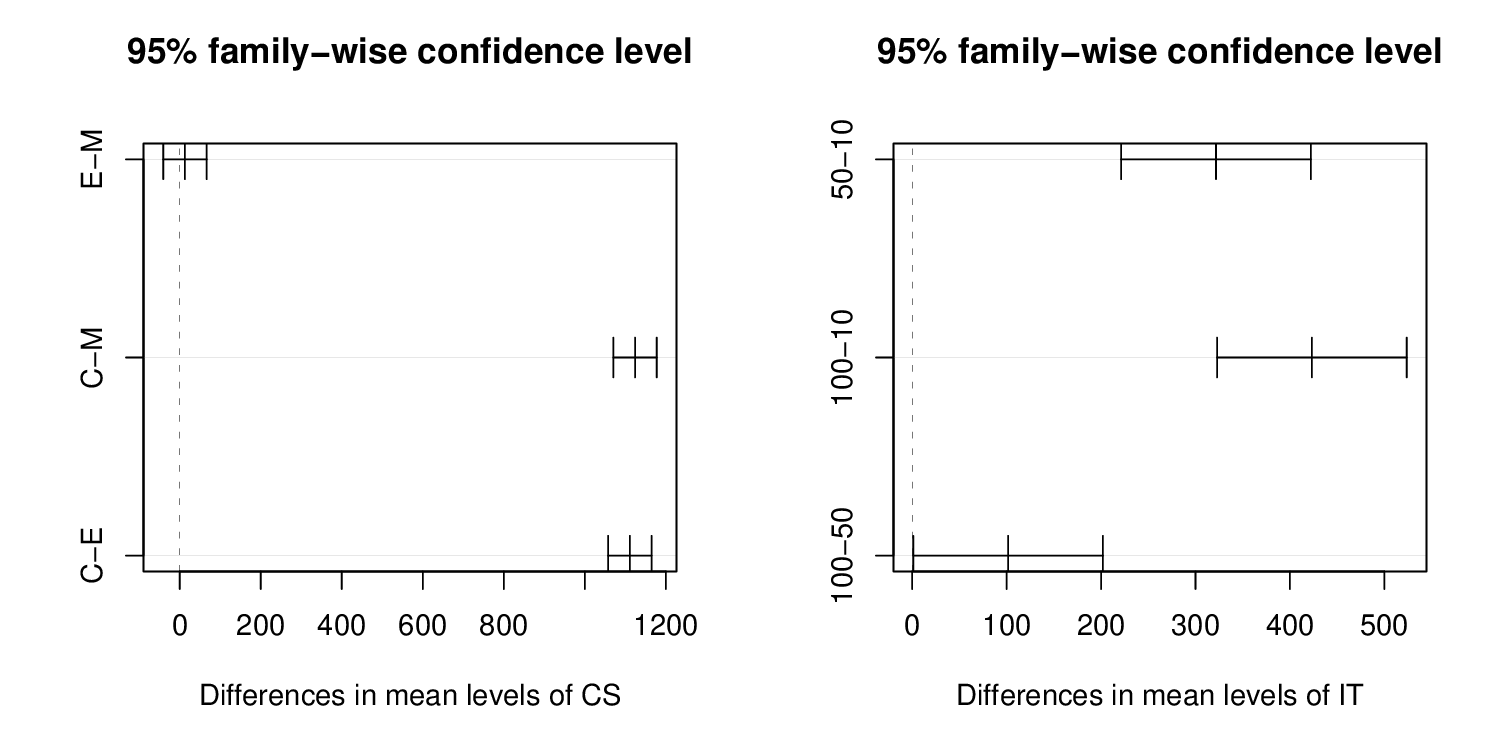}
\caption{\scriptsize{TukeyHDS test results for Griewangk function. As can be seen, no significant differences between M and E  schemes can be observed (the vertical line intersects the confidence segment corresponding to E-M). Likewise, the lowest value of \emph{IT} is the most accurate, having found significant differences when compared to the others.}}
\label{fig:tuck_sa_griewank}
\end{center}
\end{figure*}

\begin{figure*}[htp]
\begin{center}
\includegraphics[width=12cm,height=5cm]{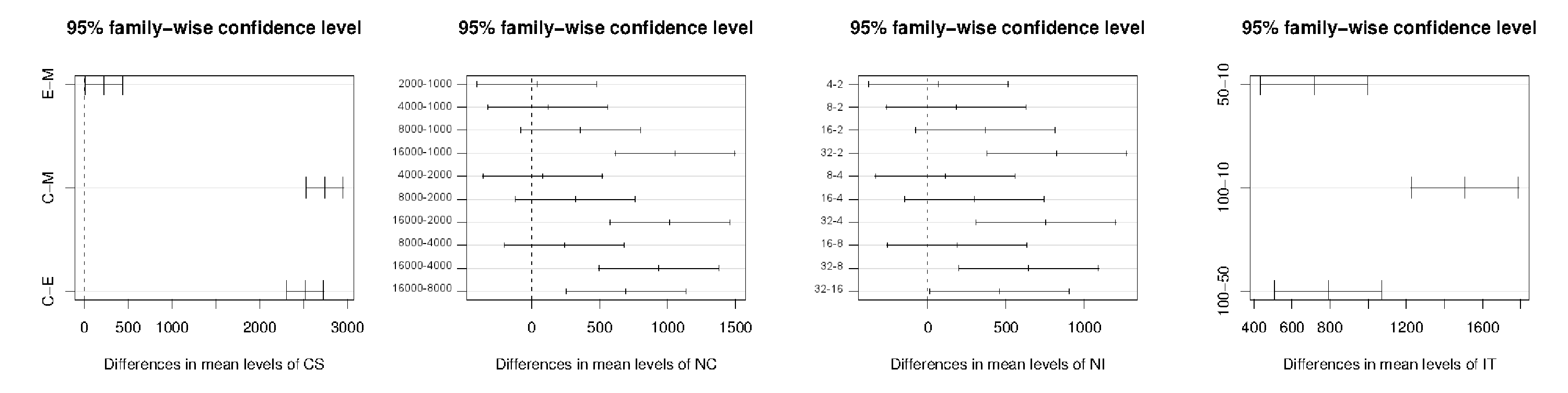}
\caption{\scriptsize{As in the previous problem, TukeyHDS test for the Rastrigin function shows no significant differences between M and E schemes. Differences between the lowest value and the highest values of \emph{IT} have been found. In the case of \emph{NC} and \emph{NI} differences appear when comparing extreme values, but not between close values (the vertical line crosses the confidence segments corresponding to comparisons between close values).}}
\label{fig:tuck_sa_rastrigin}
\end{center}
\end{figure*}

\begin{figure*}[htp]
\begin{center}
\includegraphics[width=10cm,height=5cm]{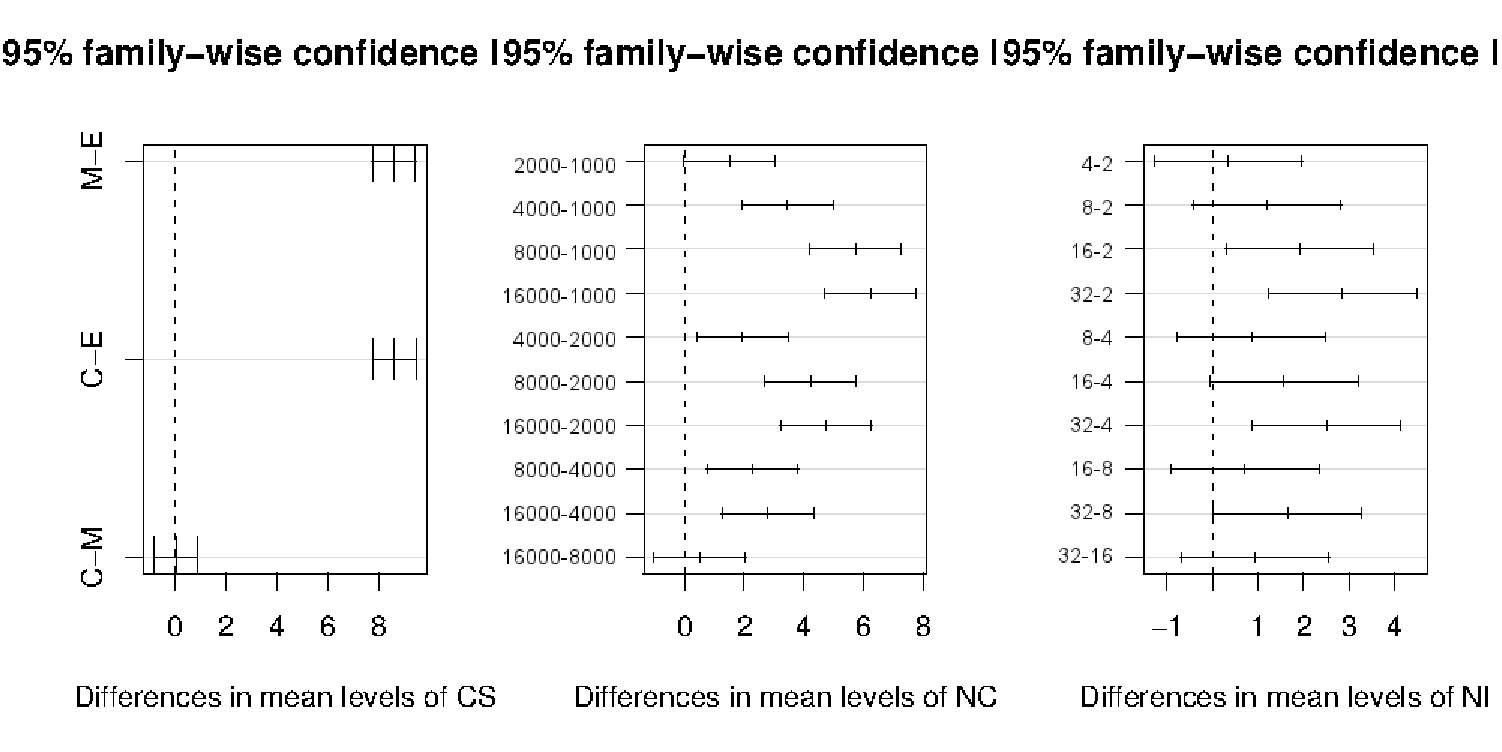}
\caption{\scriptsize{TukeyHDS test results for Ackley function show clear differences between the E and M scheme regarding C; as in previous problem, in the case of \emph{NC} and \emph{NI} differences appear when comparing extreme values.}}
\label{fig:tuck_sa_ackley}
\end{center}
\end{figure*}

\begin{figure*}[htp]
\begin{center}
\includegraphics[width=6cm,height=5cm]{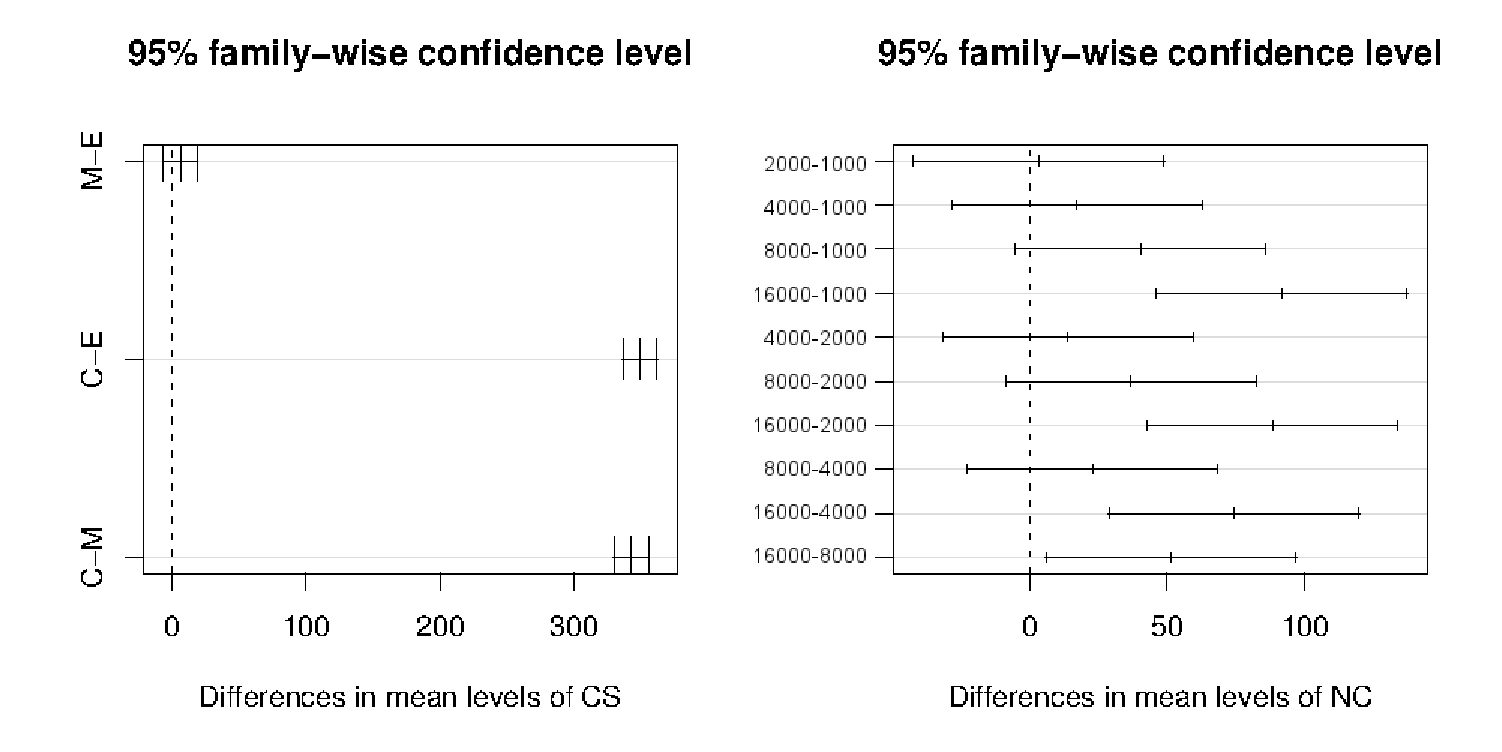}
\caption{\scriptsize{TukeyHDS test results for Rana function. As in previous functions, no significant differences between M and E schemes can be observed (although clear differences can be found regarding C). Paying attention to \emph{NC}, only small values present significant differences regarding the others (the vertical line does not intersect the confidence segments corresponding to these comparisons).}}
\label{fig:tuck_sa_rana}
\end{center}
\end{figure*}

Once the significant F-values have been obtained, Tukey HSD test will help to determine which means are significantly different. Thus, Figures \ref{fig:tuck_sa_griewank} to \ref{fig:tuck_sa_rana} show where the significant differences for those parameters with significant F-value are.

\begin{table*}[htp]

\caption{\scriptsize{Table of means for Griewangk function. As can be seen, results show significant differences in \emph{CS} and \emph{IT} cases (which is in agreement with the Tukey HSD results). Using the cooling scheme M yields to better fitness, although when compared to E scheme, differences are not significant. Finally, small values for \emph{IT} yields to better results.}
\label{Tabla:mediasGriewank_sa}}

\begin{tabular}{|c|c|c|c|c|c|} 
\hline
\scriptsize{Pa\-ra\-me\-ters}  & \multicolumn{5}{|c|}{\scriptsize{Means}}\\ 
\hline

\scriptsize{CS}   &    \multicolumn{2}{|c|}{\scriptsize{C}} &     \multicolumn{2}{|c|}{\scriptsize{M}}&  \scriptsize{E} \\
\hline 
     &\multicolumn{2}{|c|}{\scriptsize{$1155.51 $}}&  \multicolumn{2}{|c|}{\scriptsize{$ 30.66 $}} &  \scriptsize{$ 43.53 $} \\
\hline
\hline
\scriptsize{NC}   &  \scriptsize{$1000$}  &   \scriptsize{$2000$}  &   \scriptsize{$4000$}   &  \scriptsize{$8000$}   & \scriptsize{$16000$}   \\
\hline 
&\scriptsize{$460.23 $}&\scriptsize{$393.81 $}&\scriptsize{$389.94 $}&\scriptsize{$ 397.32 $}&\scriptsize{$ 408.21 $}\\ 
\hline
\hline
\scriptsize{NI}  &  \scriptsize{$2$}   &  \scriptsize{$4$}   &  \scriptsize{$8$}   & \scriptsize{$16$}    &  \scriptsize{$32$} \\ 
\hline
& \scriptsize{$424.46 $}&\scriptsize{$413.89 $}&\scriptsize{$408.19 $}&\scriptsize{$399.95 $}&\scriptsize{$403.03 $}\\
\hline
\hline
\scriptsize{PS  }   & \scriptsize{$1$}  & \scriptsize{$2$} & \scriptsize{$4$} & \scriptsize{$8$} & \scriptsize{$16$} \\ 
\hline
& \scriptsize{$456.16 $} & \scriptsize{$429.39 $} & \scriptsize{$ 406.21 $} & \scriptsize{$384.40 $} & \scriptsize{$ 373.36 $} \\ 
\hline
\hline
\scriptsize{IT}   &  \multicolumn{2}{|c|}{\scriptsize{$10$}} &   \multicolumn{2}{|c|}{\scriptsize{$50$}} &  \scriptsize{$100$} \\
\hline
 & \multicolumn{2}{|c|}{\scriptsize{$161.58 $}} & \multicolumn{2}{|c|}{\scriptsize{$483.89 $}} & \scriptsize{$584.24 $} \\
 
\hline
\end{tabular}
\end{table*}

\begin{figure*}[htp]
\begin{center}
\includegraphics[width=12cm,height=4cm]{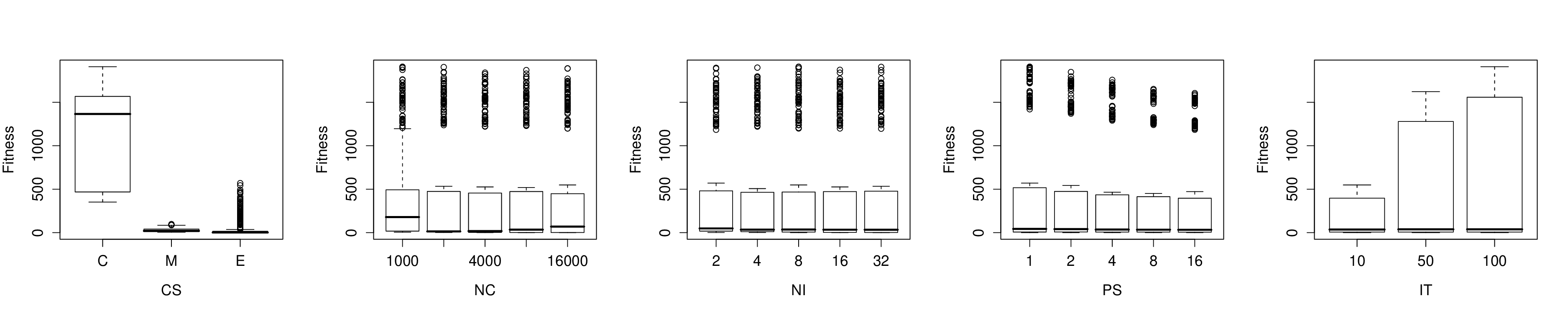}
\caption{\scriptsize{Boxplot of the set of solutions grouping using each parameter for the Griewangk function. The distribution in the case of \emph{NC, NI} and \emph{PS} is similar, while there is a clear difference with \emph{CS} and \emph{IT} (the distribution is more asymmetric and focuses on those values whose average results are better).}}
\label{fig:boxplot_sa_griewank}
\end{center}
\end{figure*}

\begin{table*}[htp]

\caption{\scriptsize{Mean fitness for Rastrigin function. As can be seen, higher values of \emph{NI} and \emph{NC} perform better on average; in the same way, using the cooling scheme M yields better results, although according to Tukey HSD, differences regarding E scheme are not significant. Finally, small values of \emph{IT} yields to better fitness.}
\label{Tabla:mediasRastrigin_sa}}

\begin{tabular}{|c|c|c|c|c|c|} 
\hline
\scriptsize{Pa\-ra\-me\-ters}  & \multicolumn{5}{|c|}{\scriptsize{Means}}\\
\hline

\scriptsize{CS}   &    \multicolumn{2}{|c|}{\scriptsize{C}} &     \multicolumn{2}{|c|}{\scriptsize{M}}&       \scriptsize{E} \\
\hline 
     &\multicolumn{2}{|c|}{\scriptsize{$3191.09 $}}&  \multicolumn{2}{|c|}{\scriptsize{$ 450.16 $}} &  \scriptsize{$ 673.93 $} \\
\hline
\hline
\scriptsize{NC }  &  \scriptsize{$1000$}  &   \scriptsize{$2000$}  &   \scriptsize{$4000$}   &  \scriptsize{$8000$}   & \scriptsize{$16000$}   \\
\hline 
&\scriptsize{$2178.65 $}&\scriptsize{$ 1483.65 $}&\scriptsize{$ 1243.50 $}&\scriptsize{$ 1162.73 $}&\scriptsize{$ 1123.42 $}\\ 
\hline
\hline
\scriptsize{NI  }&  \scriptsize{$2$}   &  \scriptsize{$   4$}   &  \scriptsize{$   8$}   & \scriptsize{$   16$}    &  \scriptsize{$32$} \\ 
\hline
& \scriptsize{$1974.32 $}&\scriptsize{$ 1516.87 $}&\scriptsize{$ 1330.77 $}&\scriptsize{$ 1219.74 $}&\scriptsize{$ 1150.25 $}\\ 
\hline
\hline
\scriptsize{PS}     &  \scriptsize{$1$}  &     \scriptsize{$2$} &     \scriptsize{$4$} &     \scriptsize{$8$} &  \scriptsize{$16$} \\ 
\hline
& \scriptsize{$1595.43 $}&\scriptsize{$ 1506.09 $}&\scriptsize{$ 1424.15 $}&\scriptsize{$ 1349.61 $}&\scriptsize{$ 1316.68 $}\\ 
\hline
\hline
\scriptsize{IT}   &  \multicolumn{2}{|c|}{\scriptsize{$10$}} &   \multicolumn{2}{|c|}{\scriptsize{$50$}} &    \scriptsize{$100$} \\
\hline
     &  \multicolumn{2}{|c|}{\scriptsize{$697.43 $}} & \multicolumn{2}{|c|}{\scriptsize{$1413.62 $}} & \scriptsize{$2204.12 $} \\
 
\hline
\end{tabular}
\end{table*} 

\begin{figure*}[htp]
\begin{center}
\includegraphics[width=12cm,height=4cm]{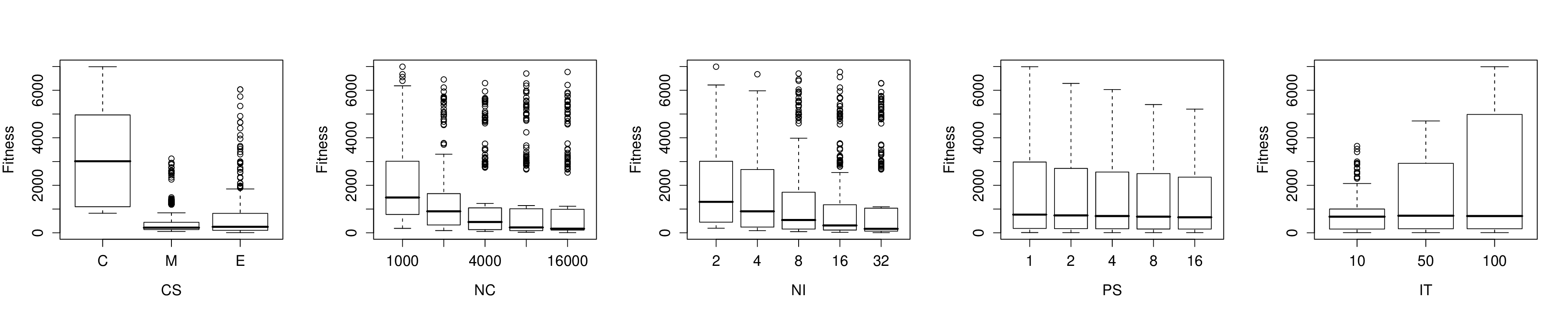}
\caption{\scriptsize{The boxplot of the set of solutions for the Rastrigin function shows differences for \emph{CS} and \emph{IT}. The box corresponding to \emph{PS} shows a homogeneous graph, which indicates that no parameter values differ from the rest.}}
\label{fig:boxplot_sa_rastrigin}
\end{center}
\end{figure*}

\begin{table*}[htp]

\caption{\scriptsize{Table of means for Ackley function. In this problem, best fitness are obtained using higher values of \emph{NI} and \emph{NC}, while taking into account \emph{CS}, using the E scheme yields to best results (significant differences regarding other schemes can be found). Paying attention to \emph{IT} and \emph{PS}, no significant differences have been found.}
\label{Tabla:mediasAck_sa}}

\begin{tabular}{|c|c|c|c|c|c|} 
\hline
\scriptsize{Pa\-ra\-me\-ters}  & \multicolumn{5}{|c|}{\scriptsize{Means}}\\
\hline

\scriptsize{CS}   &    \multicolumn{2}{|c|}{\scriptsize{C}} &     \multicolumn{2}{|c|}{\scriptsize{M}}&       \scriptsize{E }\\
\hline 
     &\multicolumn{2}{|c|}{\scriptsize{$21.65 $}}&  \multicolumn{2}{|c|}{\scriptsize{$ 21.62 $}} &  \scriptsize{$ 13.053 $ }\\
\hline
\hline
\scriptsize{NC   }&  \scriptsize{$1000$  }&   \scriptsize{$2000$  }&   \scriptsize{$4000$   }&  \scriptsize{$8000$   }& \scriptsize{$16000$   }\\
\hline 
&\scriptsize{$21.63 $}&\scriptsize{$ 21.12 $}&\scriptsize{$18.84 $}&\scriptsize{$ 16.89 $}&\scriptsize{$15.39 $}\\ 
\hline
\hline
\scriptsize{NI  }&       \scriptsize{$2$   }&  \scriptsize{$   4$   }&  \scriptsize{$   8$   }& \scriptsize{$   16$    }&  \scriptsize{$32$} \\ 
\hline
&\scriptsize{$20.36 $}&\scriptsize{$ 19.43 $}&\scriptsize{$18.71 $}&\scriptsize{$17.86 $}&\scriptsize{$17.51 $}\\
\hline
\hline
\scriptsize{PS     }&  \scriptsize{$1$  }&     \scriptsize{$2$}&     \scriptsize{$4$} &     \scriptsize{$8$} &       \scriptsize{$16$} \\ 
\hline
&\scriptsize{$18.98 $}&\scriptsize{$ 18.84 $}&\scriptsize{$18.78 $}&\scriptsize{$ 18.67 $}&\scriptsize{$ 18.61 $} \\ 
\hline
\hline
\scriptsize{IT   }&  \multicolumn{2}{|c|}{\scriptsize{$10$}} &   \multicolumn{2}{|c|}{\scriptsize{$50$}} &    \scriptsize{$100$} \\
\hline
     &  \multicolumn{2}{|c|}{\scriptsize{$18.74 $}} & \multicolumn{2}{|c|}{\scriptsize{$18.78 $}} & \scriptsize{$18.80 $} \\

\hline
\end{tabular}
\end{table*} 

\begin{figure*}[htp]
\begin{center}
\includegraphics[width=12cm,height=4cm]{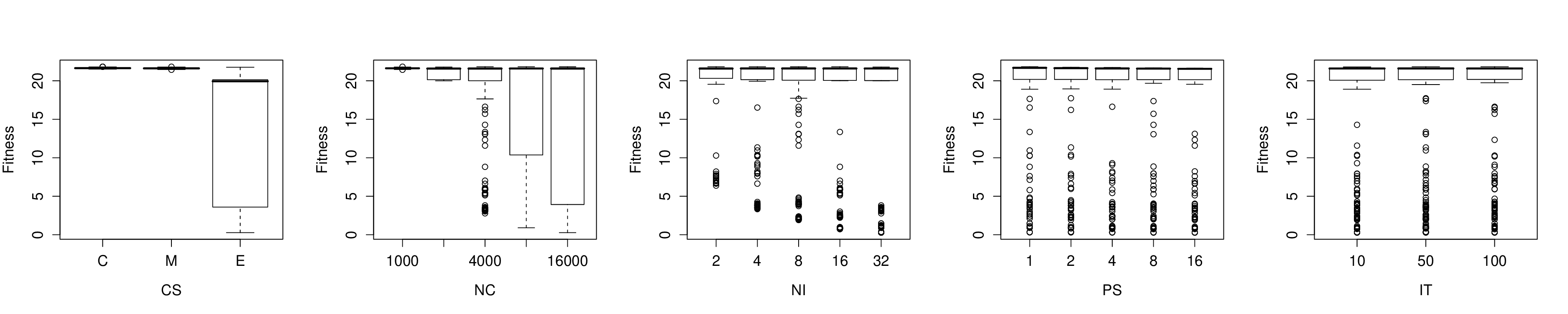}
\caption{\scriptsize{Boxplot for Ackley function. There are many cases outside the quartiles represented (due to the characteristics of the problem.) Taking into account \emph{CS}, \emph{CN} and \emph{NI}, differences between different groups can be found; while for \emph{IT} and \emph{PS} no differences can be observed.}}
\label{fig:boxplot_sa_ackley}
\end{center}
\end{figure*}

\begin{table*}[htp]

\caption{\scriptsize{Mean fitness for Rana function. According to ANOVA, \emph{CS} and \emph{NC} parameters are statistically significant. Taking into account the Tukey HSD results too, it can be claimed that the best value for \emph{CS} is the E scheme (although no differences regarding M have been found). As far as \emph{NC} parameter is concerned, a middle value yields the best results, while for \emph{NI, PS} and \emph{IT}, no significant differences between the values tested can be found.}
\label{Tabla:mediasRana_sa}}

\begin{tabular}{|c|c|c|c|c|c|} 
\hline
\scriptsize{Pa\-ra\-me\-ters}  & \multicolumn{5}{|c|}{\scriptsize{Means}}\\
\hline

\scriptsize{CS   }&    \multicolumn{2}{|c|}{\scriptsize{C}} &     \multicolumn{2}{|c|}{\scriptsize{M}}&       \scriptsize{E }\\
\hline 
 & \multicolumn{2}{|c|}{\scriptsize{$474.30 $}} & \multicolumn{2}{|c|}{\scriptsize{$131.69 $}}&\scriptsize{$ 125.24 $}\\ 
\hline
\hline
\scriptsize{NC   }&  \scriptsize{$1000$  }&   \scriptsize{$2000$  }&   \scriptsize{$4000$   }&  \scriptsize{$8000$   }& \scriptsize{$16000$   }\\
\hline 
&\scriptsize{$304.83 $}&\scriptsize{$230.59 $}&\scriptsize{$213.24 $}&\scriptsize{$216.60 $}&\scriptsize{$253.46 $}\\
\hline
\hline
\scriptsize{NI  }&       \scriptsize{$2$   }&  \scriptsize{$   4$   }&  \scriptsize{$   8$   }& \scriptsize{$   16$    }&  \scriptsize{$32$} \\ 
\hline
&\scriptsize{$268.33 $}&\scriptsize{$253.37 $}&\scriptsize{$240.22 $}&\scriptsize{$ 231.80 $}&\scriptsize{$225.00 $} \\ 
\hline
\hline
\scriptsize{PS     }&  \scriptsize{$1$  }&     \scriptsize{$2$} &     \scriptsize{$4$} &     \scriptsize{$8$} &       \scriptsize{$16$} \\ 
\hline
&\scriptsize{$255.99 $}&\scriptsize{$ 248.53 $}&\scriptsize{$ 242.45 $}&\scriptsize{$238.36 $}&\scriptsize{$233.39 $}\\
\hline
\hline
\scriptsize{IT   }&  \multicolumn{2}{|c|}{\scriptsize{$10$}} &   \multicolumn{2}{|c|}{\scriptsize{$50$}} &    \scriptsize{$100$} \\
\hline
 & \multicolumn{2}{|c|}{\scriptsize{$228.055 $}} & \multicolumn{2}{|c|}{\scriptsize{$248.47 $}}&\scriptsize{$254.71 $}\\
 
\hline
\end{tabular}
\end{table*}

\begin{figure*}[htp]
\begin{center}
\includegraphics[width=12cm,height=4cm]{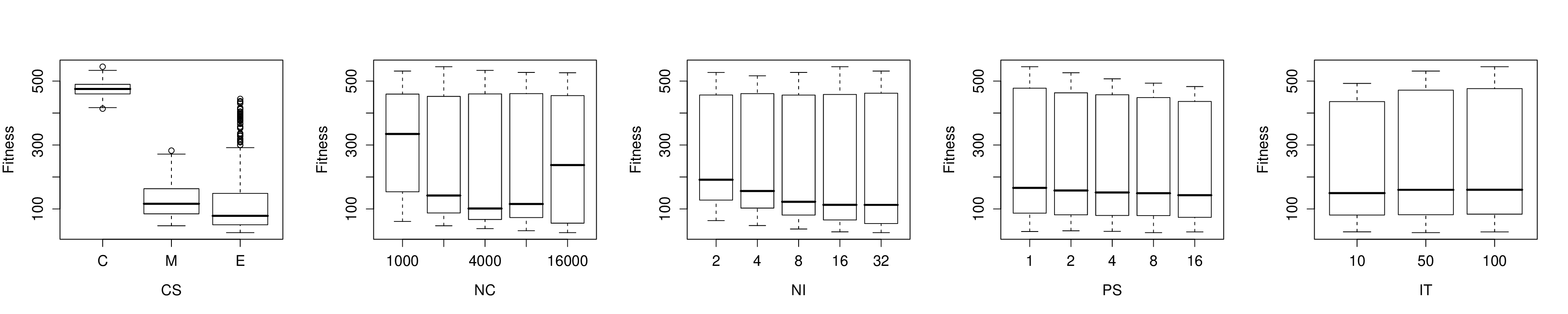}
\caption{\scriptsize{The boxplot of the set of solutions for the Rana function shows the results discussed in Table \ref{Tabla:mediasRana_sa}. Taking into account \emph{CS},  M and E schemes are very similar, while for \emph{NC} significative differences between distant parameter values can be found (although no differences between close values).}}
\label{fig:boxplot_sa_rana}
\end{center}
\end{figure*}

Tukey HSD test shows that in general, no significative differences can be found between M or E schemes (the vertical line intersects the confidence segment corresponding to E-M), although clear differences appear regarding C (it is the least effective).
\emph{PS} parameter has not been tested with Tukey HSD as it has not been reported significant according to ANOVA test.
In general, the initial temperature (\emph{IT}) must be kept low, value for which significant differences were found.
For the rest of parameters (\emph{NC} and \emph{NI}), in those problems where these parameters were found as significant according to ANOVA, significative differences between pairs of extreme values do exist (the vertical line -distance 0- does not intersect those confidence segments corresponding to comparisons of extreme values).

After the parameters with greater influence on the results have been determined, accurate parameter values should be established in order to obtain an optimal operation. 
To do so, tables of means and boxplots \cite{boxplot1988} are obtained to show the effect each level has on the approximation error (see Tables \ref{Tabla:mediasGriewank_sa} to \ref{Tabla:mediasRana_sa} and Figures \ref{fig:boxplot_sa_griewank} to \ref{fig:boxplot_sa_rana}).

Paying attention to the cooling schedule, in general there are no differences between M and E schedules; however, clear differences can be found regarding C schedule.
As far as \emph{NC} and \emph{NI} are concerned, the higher these parameter values, the better the average fitness  (as these parameters determine how many times the algorithm tries to find a new solution).
Using small values for \emph{IT} leads to obtain better average results.
Taking into account \emph{PS}, no clear differences can be found (according to the results of ANOVA and Tukey HSD tests).

The analysis shows that obtained results are in agreement with those obtained by means of theoretical analysis and available in bibliography \cite{devicente2003,Weyland2008}. 
Moreover, obtained results might help practitioners in adjusting parameters of new optimization methods.

\subsection{Experiment 2: Parameter estimate in the lognormal diffusion process}
\label{exp2}

Next, the ANOVA and Tukey HSD statistical tools are applied to determine the influence of parameter values on the obtained fitness for the likelihood function. 
ANOVA tables are shown as well as Tukey HSD plots and tables of means.

\begin{table}[htp]
\caption{\scriptsize{Likelihood function. ANOVA table for the fitness (response) with the optimization algorithm parameters as factors.}
\label{tablas:anova:like}}
    \begin{tabular}{|c|c|c|c|}
    \hline
    \scriptsize{Param.}        &\scriptsize{FD}&  \scriptsize{F} & \scriptsize{$P (Sig. Level)$} \\
    \hline
    \scriptsize{CS} 		&\scriptsize{2}&\scriptsize{$ 0.151  $}&\scriptsize{$ 0.860 $}\\ 
    \scriptsize{\textbf{NC}} 	&\scriptsize{4}&\scriptsize{$ 125.07 $}&\scriptsize{\textbf{$ <2.2e-16 $}}\\ 
    \scriptsize{\textbf{NI}} 	&\scriptsize{4}&\scriptsize{$ 124.43  $}&\scriptsize{\textbf{$  <2.2e-16$}}\\
    \scriptsize{\textbf{PS}}    &\scriptsize{4}&\scriptsize{$ 4.23  $}&\scriptsize{\textbf{$0.002 $}}\\
    \scriptsize{IT} 		&\scriptsize{2}&\scriptsize{$  0.43 $}&\scriptsize{$0.649 $}\\ 
    \hline
    \end{tabular}
\end{table}

Table \ref{tablas:anova:like} shows the result of applying ANOVA on likelihood function using the SQ algorithm. 
Parameters with a significance level over $95\%$ are highlighted in boldface.

\begin{figure*}[htp]
\begin{center}
\includegraphics[width=10cm,height=5cm]{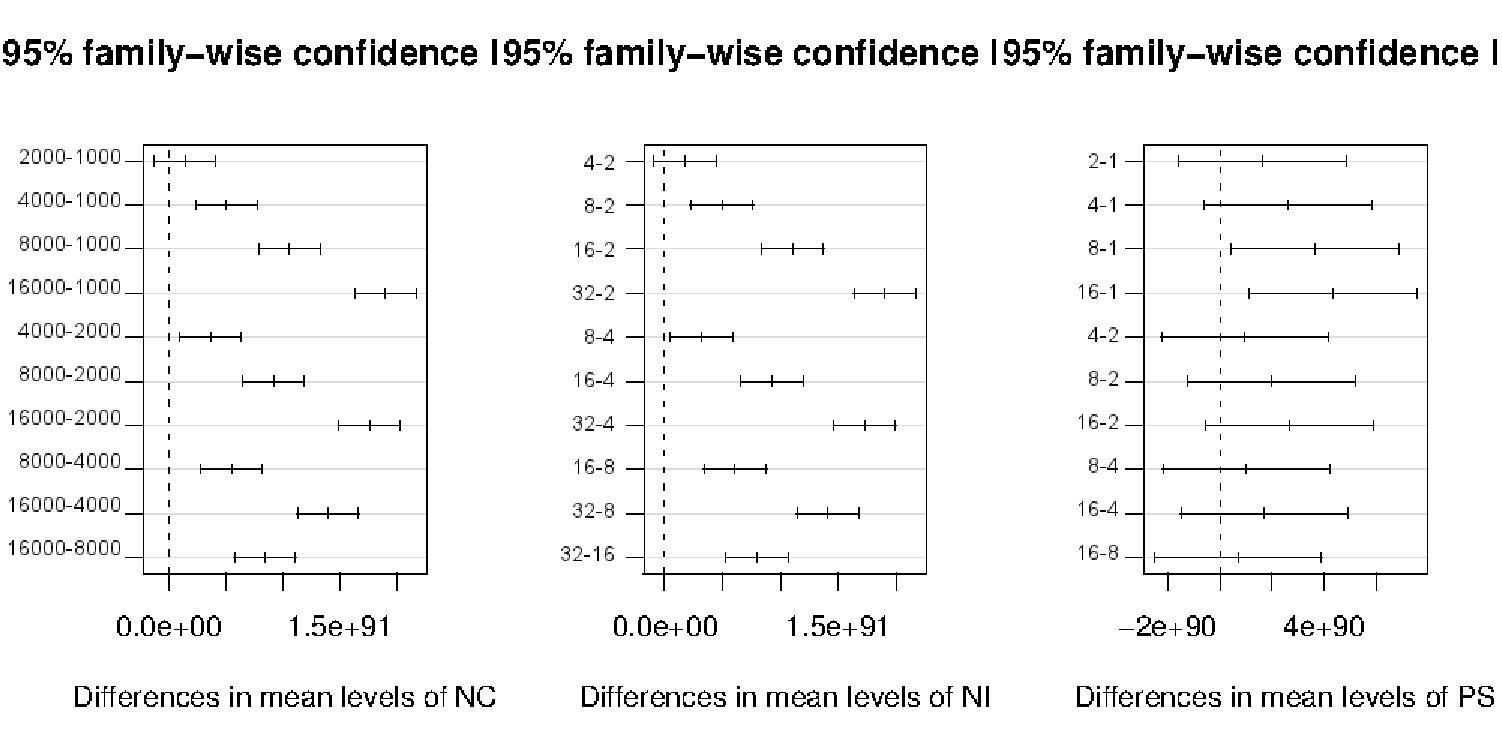}
\caption{\scriptsize{TukeyHDS test results for Likelihood function. In the case of \emph{NC}, \emph{NI} and \emph{PS} differences appear when comparing extreme values, but not between close values (the vertical line crosses the confidence segments corresponding to comparisons between close values).}}
\label{fig:tuck_sa_like}
\end{center}
\end{figure*}

    \begin{table*}[htp]

    \caption{\scriptsize{Mean fitness for Likelihood function. As can be seen, results show significant differences in \emph{NI}, \emph{NC} and \emph{PS} cases (which is in agreement with the ANOVA and Tukey HSD results). Taking into account \emph{CS}, using the E scheme yields to best results (although significant differences regarding other schemes could not be found). Paying attention to \emph{IT}, small values of \emph{IT} yields to better fitness.}
    \label{tabla:medias_sa_like}}

    \begin{tabular}{|c|c|c|c|c|c|} 
    \hline
    \scriptsize{Pa\-ra\-me\-ters}  & \multicolumn{5}{|c|}{\scriptsize{Means}}\\
    \hline
    \scriptsize{CS}   &    \multicolumn{2}{|c|}{\scriptsize{C}} &     \multicolumn{2}{|c|}{\scriptsize{M}}&       \scriptsize{E} \\
    \hline
    &\multicolumn{2}{|c|}{\scriptsize{$8.16e+90 $}}&  \multicolumn{2}{|c|}{\scriptsize{$ 7.65e+90 $}} &  \scriptsize{$ 7.94e+90 $} \\

    \hline    \hline
    \scriptsize{NC }  &  \scriptsize{$1000$}  &   \scriptsize{$2000$}  &   \scriptsize{$4000$}   &  \scriptsize{$8000$}   & \scriptsize{$16000$}   \\
    \hline
    &\scriptsize{$6.96e+89 $}&\scriptsize{$ 2.11e+90 $}&\scriptsize{$ 5.77e+90 $}&\scriptsize{$ 1.13e+91 $}&\scriptsize{$ 1.97e+91 $}\\
    \hline    \hline
    \scriptsize{NI  }&  \scriptsize{$2$}   &  \scriptsize{$   4$}   &  \scriptsize{$   8$}   & \scriptsize{$   16$}    &  \scriptsize{$32$} \\
    \hline
    & \scriptsize{$6.65e+89 $}&\scriptsize{$ 2.35e+90 $}&\scriptsize{$ 5.50e+90 $}&\scriptsize{$ 1.16e+91 $}&\scriptsize{$ 1.96e+91 $}\\
    \hline    \hline
    \scriptsize{PS}     &  \scriptsize{$1$}  &     \scriptsize{$2$} &     \scriptsize{$4$} &     \scriptsize{$8$} &  \scriptsize{$16$} \\
    \hline
    & \scriptsize{$5.47e+90 $}&\scriptsize{$7.12e+90 $}&\scriptsize{$8.09e+90 $}&\scriptsize{$ 9.11e+90 $}&\scriptsize{$ 9.80e+91 $}\\
    \hline    \hline
    \scriptsize{IT}   &  \multicolumn{2}{|c|}{\scriptsize{$10$}} &   \multicolumn{2}{|c|}{\scriptsize{$50$}} &    \scriptsize{$100$} \\
    \hline
    &  \multicolumn{2}{|c|}{\scriptsize{$8.37e+90 $}} & \multicolumn{2}{|c|}{\scriptsize{$7.53e+90 $}} & \scriptsize{$7.86e+90 $} \\
    \hline
    \end{tabular}
    \end{table*}

\begin{figure*}[htp]
\begin{center}
\includegraphics[width=12cm,height=4cm]{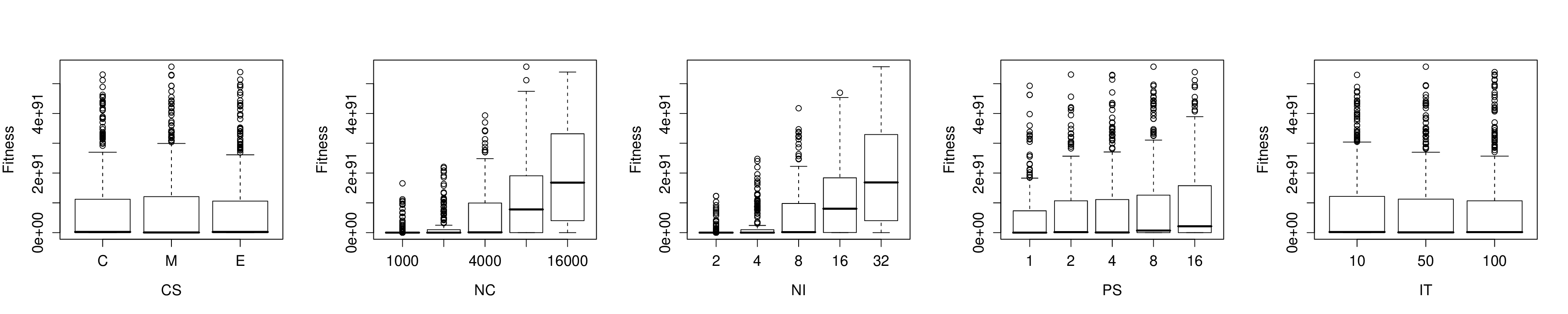}
\caption{\scriptsize{The boxplot of the set of solutions for the Likelihood function shows differences for \emph{NC}, \emph{NI} and \emph{PS} (significative differences between distant parameter values can be found (although no differences between close values). Boxes corresponding to \emph{CS} and \emph{IT} show homogeneous graphs, which indicates that no parameter values differ from the rest.}}
\label{fig:boxplot_sa_like}
\end{center}
\end{figure*}

The ANOVA analysis shows that NC and NI parameters influence the obtained fitness. In this case, PS has been reported as significant; it might be due to the difficulty of this problem in which using several states help to explore the search space.

Again, once the significant F-values have been obtained, Tukey HSD test help to determine which means are significantly different. Thus, Figure \ref{fig:tuck_sa_like} shows where the significant differences for those parameters with significant F-value are.

Tukey HSD shows that NC, NI and PS are significant when pairs of extreme values are compared.
CS and IT parameters have not been tested as they have not been reported significant according to ANOVA; in any case, these parameters will be examined through tables of means and boxplots.

After the parameters with greater influence on the results have been determined, tables of means and boxplots have been obtained to show the effect each level has on the approximation error (see Table \ref{tabla:medias_sa_like} and Figure \ref{fig:boxplot_sa_like}).

As can be seen, higher values of \emph{NI}, \emph{NC} and \emph{PS} perform better on average.
In the same way, using the cooling scheme E yields better results, although according to ANOVA, this parameter was not reported as significant. 
Finally, small values of \emph{IT} yields to better fitness, although no significant differences have been found.

In any case, in this paper we have verified the relative importance of these parameters on the fitness through a rigorous statistical study.

\section{Conclusions and Work in Progress}
\label{sec:conclusionsAndFutureWork}

This paper proposes a methodology to analyze the parameters with a higher influence on the performance for a given optimization method.
Our approach has been tested using a simulated annealing based algorithm (SQ), a widespread optimization method successfully used by practitioners in many areas. 
We also report appropriate values (among those tested) for these parameters in order to obtain an optimal operation.

A statistical study of the different parameters involved in the design of this optimization method has been carried out by applying ANOVA, which consists of a set of statistical techniques that analyze and compare experiments by describing the interactions and interrelations between the variables or factors of the system, and completed with a Tukey HSD test. 
The motivation of the present statistical study lies in the great variety of alternatives that a designer has to take into account when designing an optimization method.

The proposed methodology has been applied to four well-known function approximation problems (Griewangk, Rastrigin, Ackley and Rana), widely used by practitioners, in order to determine which parameters have a higher influence on results (a change on those parameters will affect the algorithm performance). 
In a second experiment, a much complex problem has been addressed. In this case, the estimation of the parameters involved in a stochastic model (lognormal diffusion process \cite{Gutierrez2001}) has been carried out.

\medskip

In that context, results show that in most cases, using the ''Exponential'' and ''Modified Cauchy'' cooling schemes yields better fitness values. 
If the initial solution is good enough, a cooling scheme where the probability of accepting a worst solution is higher, is more appropriate to achieve a better solution (i.e. ''Exponential''). However, if the initial solution is not good enough, then a cooling scheme that allows accepting a worse solution is more accurate. In this sense, ''Modified Cauchy'' works better.
However, according to the Tukey HSD test results, no significative differences can be found between them, although clear differences can be found with ''Cauchy'' schedule.
In general, accurate results are obtained for high values of \emph{NC} and \emph{NI} parameters (the higher these parameter values, the better the average fitness obtained, as these parameters determine how many times the algorithm tries to find a new solution). 
Taking into account the Tukey HSD test, only for extreme values (far one from the others) significative differences can be found.
As far as the \emph{IT} parameter is concerned, using small values for this parameter leads to obtain better average results. 
The Tukey HSD test confirms this fact showing that small values present significative differences regarding higher values.
Paying attention to \emph{PS}, using a high value yields to better fitness values, although this increases the number of evaluations and time needed to run the algorithm. However it has not been reported as significant, which is in agreement to the use of the typical SA algorithm (that uses a single individual or state).

In the case of the lognormal diffusion process parameter estimate, results show that 
using the E scheme yields to best results (although significant differences regarding other schemes could not be found).
Accurate results are obtained using a high value of \emph{PS}. This might be due to the difficulty of the problem (using several states help to explore the search space).
As in the previous problems, using a high value of \emph{NC} and \emph{NI} parameters yields to accurate results (as these parameters determine how many times the algorithm tries to find a new solution, and this results on a wider exploration). 
Finally, small values of \emph{IT} yields to better fitness, although no significant differences have been found according to ANOVA and Tukey HSD results.

\medskip

As can be seen, different algorithms or configurations might work better on a problem and worse on another one.
This is in agreement with the \textit{No Free Lunch} theorem \cite{Wolpert1997}, according to which there is no algorithm better than all to solve all the problems.

This methodology based on ANOVA and TukeyHDS statistical tests could be helpful for practitioners in analyzing and adjusting parameters of any optimization method.

Our work in progress includes the analysis of different optimization methods, such as modified EAs considering several selection schemes, new genetic operators and other meta-heuristics (tabu search, EDAs, UMDA, etc).
As future work, the implementation of a parameter control method would be of interest, as proposed by Eiben et al. \cite{emss2007}. In this case, ANOVA could be used to analyze not only the optimization method parameters but also the control strategy parameters.

Finally, software is available under GNU public license at {\tt http://genmagic.ugr.es}

\section*{acknowledgements}
This work has been supported in part by 
the CEI BioTIC GENIL (CEB09-0010) MICINN CEI Program (PYR-2010-13) project, 
the Junta de Andaluc\'{\i}a TIC-3903 and P08-TIC-03928 projects, and 
the Ja\'en University UJA-08-16-30 project.
The authors are very grateful to the anonymous referees whose comments and suggestions have contributed to improve this paper.



\section*{Appendix I}
\label{sec_apendice_r}
This appendix shows an exhaustive review of commands to R used to analyze data obtained after the runs and to obtain tables and figures shown in Section \ref{sec:Analysis}. As an example, those commands corresponding to the Ackley function are shown:

\begin{itemize}
\item \scriptsize{  problemData $<$- Ackley  }
\item \scriptsize{  problemName $<$- "Ackley"  }
\item \scriptsize{  pathMeans $<$- "/APIN/R/Medias"  }
\item \scriptsize{  pathTukeyHDS $<$- "/APIN/R/Tukey"  }
\item \scriptsize{  pathBoxPlot $<$- "/APIN/R/BoxPlot"  }
\item \scriptsize{  extension $<$- ".eps"  }
\item \scriptsize{  dirTukeyHDS $<$- paste(pathTukeyHDS,problemName,sep = "")  }
\item \scriptsize{  dirMeans $<$-paste(pathMeans,problemName,sep = "")  }
\item \scriptsize{  dirBox $<$-paste(pathBoxPlot,problemName,sep = "")  }
\item \scriptsize{  dirTukeyHDS $<$- paste(dirTukeyHDS,extension,sep = "")  }
\item \scriptsize{  dirMeans $<$-paste(dirMeans,extension,sep = "")  }
\item \scriptsize{  dirBox $<$-paste(dirBox,extension,sep = "")  }

\item \scriptsize{  problemData\$NC $<$- as.factor(problemData\$NC)  }
\item \scriptsize{  problemData\$NI $<$- as.factor(problemData\$NI)  }
\item \scriptsize{  problemData\$PS $<$- as.factor(problemData\$PS)  }
\item \scriptsize{  problemData\$IT $<$- as.factor(problemData\$IT)  }
\item \scriptsize{  problemData\$CS $<$- factor(problemData\$CS, labels=c('C','M','E'))  }

\item \scriptsize{  shapiro.test(problemData\$Fitness)  }
\item \scriptsize{  levene.test(problemData\$Fitness, problemData\$NC)  }
\item \scriptsize{  levene.test(problemData\$Fitness, problemData\$NI)  }
\item \scriptsize{  levene.test(problemData\$Fitness, problemData\$PS)  }
\item \scriptsize{  levene.test(problemData\$Fitness, problemData\$IT)  }
\item \scriptsize{  levene.test(problemData\$Fitness, problemData\$CS)  }

\item \scriptsize{  summary(anovaNC$<$-(aov(Fitness ~ NC, data=problemData)))  }
\item \scriptsize{  summary(anovaNI$<$-(aov(Fitness ~ NI, data=problemData)))  }
\item \scriptsize{  summary(anovaPS$<$-(aov(Fitness ~ PS, data=problemData)))  }
\item \scriptsize{  summary(anovaIT$<$-(aov(Fitness ~ IT, data=problemData)))  }
\item \scriptsize{  summary(anovaCS$<$-(aov(Fitness ~ CS, data=problemData)))  }

\item \scriptsize{  tapply(problemData\$Fitness, list(problemData\$NC), mean)  }
\item \scriptsize{  tapply(problemData\$Fitness, list(problemData\$NI), mean)  }
\item \scriptsize{  tapply(problemData\$Fitness, list(problemData\$PS), mean)  }
\item \scriptsize{  tapply(problemData\$Fitness, list(problemData\$IT), mean)  }
\item \scriptsize{  tapply(problemData\$Fitness, list(problemData\$CS), mean)  }

\item \scriptsize{  layout(matrix(1:4,1,4))  }
\item \scriptsize{  plot(Tukey HSD(anovaNC, "NC"))  }
\item \scriptsize{  plot(Tukey HSD(anovaNI, "NI"))  }
\item \scriptsize{  plot(Tukey HSD(anovaPS, "PS"))  }
\item \scriptsize{  plot(Tukey HSD(anovaIT, "IT"))  }
\item \scriptsize{  plot(Tukey HSD(anovaCS, "CS"))  }
\item \scriptsize{  dev.copy2eps(file=dirTukeyHDS, width=12.0, height=12.0, pointsize=12)  }

\item \scriptsize{  layout(matrix(1:4,1,4))  }
\item \scriptsize{  boxplot(Fitness~NC, ylab="Fitness", xlab="NC", data=problemData)  }
\item \scriptsize{  boxplot(Fitness~NI, ylab="Fitness", xlab="NI", data=problemData)  }
\item \scriptsize{  boxplot(Fitness~PS, ylab="Fitness", xlab="PS", data=problemData)  }
\item \scriptsize{  boxplot(Fitness~IT, ylab="Fitness", xlab="IT", data=problemData)  }
\item \scriptsize{  boxplot(Fitness~CS, ylab="Fitness", xlab="CS", data=problemData)  }

\item \scriptsize{  dev.copy2eps(file=dirBox, width=12.0, height=12.0, pointsize=12)  }

\end{itemize}

\end{document}